# An Intelligent Monitoring System of Vehicles on Highway Traffic

Sulaiman Khan[1,2], Hazrat Ali[3*], Zia Ullah[1] and Mohammad Farhad Bulbul[4]

[1] Department of Computer Systems Engineering, University of Engineering and Technology Peshawar, Pakistan.
[2] Department of Computer Science, University of Swabi, Pakistan.
[3] Department of Electrical and Computer Engineering, COMSATS University Islamabad, Abbottabad Campus, Abbottabad, Pakistan.
[4] Department of Mathematics, Jessore University of Science and Technology, Bangladesh.
Email:engr.hazratali@yahoo.com, hazratali@ciit.net.pk,

*Abstract*— Vehicle speed monitoring and management of highways is the critical problem of the road in this modern age of growing technology and population. A poor management results in frequent traffic jam, traffic rules violation and fatal road accidents. Using traditional techniques of RADAR, LIDAR and LASAR to address this problem is time-consuming, expensive and tedious. This paper presents an efficient framework to produce a simple, cost efficient and intelligent system for vehicle speed monitoring. The proposed method uses an HD (High Definition) camera mounted on the road side either on a pole or on a traffic signal for recording video frames. On the basis of these frames, a vehicle can be tracked by using radius growing method, and its speed can be calculated by calculating vehicle mask and its displacement in consecutive frames. The method uses pattern recognition, digital image processing and mathematical techniques for vehicle detection, tracking and speed calculation. The validity of the proposed model is proved by testing it on different highways.

## I. Introduction

Traffic increase is an ever growing phenomenon. The traffic rise directly creates challenges for robust traffic flow, not only in urban areas but also in rural areas. Road accidents are a direct consequence of uncontrolled traffic flow and poorly monitored speed. And road accidents lead to loss of lives. Thus, detecting a particular vehicle and monitoring its speed through machines is important. Previously, the approaches used for this task were based on RADAR and LIDAR. Sawicki et al., [1] presented the used of RADAR for vehicle speed calculation. A RADAR works on the principle of bouncing a radio signal back from a moving object and the receiver receives the reflected signal. The receiver RADAR calculates vehicle speed on the basis of calculated difference in frequency. Given the high cost and susceptibility to target identification error, RADAR has not received widespread implementation for vehicles speed monitoring. On the other hand, a LIDAR device works on the principle of generating a light pulse from a LIDAR gun and calculating the time taken by the pulse to travel to the moving object and return to the LIDAR gun. With the help of this information, a LIDAR device can calculate the distance between a moving object and a LIDAR gun. Through a recursive process, a LIDAR can calculate moving object speed accurately. However, LIDAR also incurs high cost. The (LASER) has also been under study however, it also has high cost and suffers incorrect results in fog, rain and snow. Abbas et al., [2] presented the approach of using license number plate for speed measurement, however this approach has a limitation in the situation when two vehicles appear together after each other in a frame, then it is impossible to detect the number plate of the second vehicle. Singh et al., [4] presented an approach of Radio Frequency Identification (RFID) technology for detecting a speed violation. Sina et al.,[5] presented the approach of headlight detection to monitor a moving vehicle and then, determined the speed by pin hole and Euclidian distance methods, however this method also has limitations if two vehicles appear in the same frame, as detection of the rear vehicle becomes challenging. Song et al.,[6] presented a genetic programming approach for detecting vehicle speed. Lin et al., [7] used an approach of imaging geometry, blur extent in the image and camera pose to calculate vehicle speed. Rad et al., [8] used an approach of computer vision techniques for speed measurement. These approaches suffer issues such as number plate occlusion, headlight occlusion, and/or being costly.

In order to address these problems, in this work, a cost effective and efficient model is presented for speed monitoring of vehicles on highways. In fact, the model is based on a high definition (HD) camera connected to a computer to detect over-speeding. The camera is fixed on road side or on traffic signal pole for video stream recording and the computer performs logical operations, morphological operations and speed calculations for each vehicle. The modules of the proposed model are shown in Fig. 1.

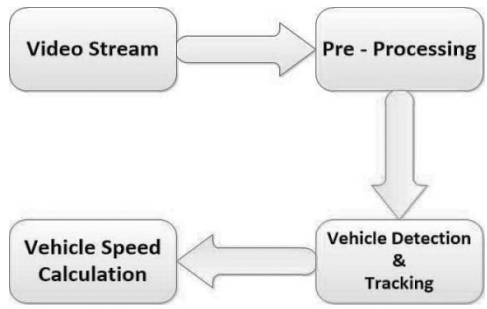

*Figure 1. Modules of proposed system*

## II. PROPOSED FRAMEWORK

This research work presents an efficient framework for efficient traffic surveillance. The flowchart of our proposed method is depicted in Fig. 1 and all the steps in the figure are discussed in the following text in detail. Notice that the flowchart is constructed through the sub-modules of modules illustrated in Fig. 2. More clearly, this section discusses all the sub-modules of modules represented in Fig. 1. For example, the pre-processing step consists of convert video into frames, background elimination, lane masking, noise removing, object contours linking and labelling.

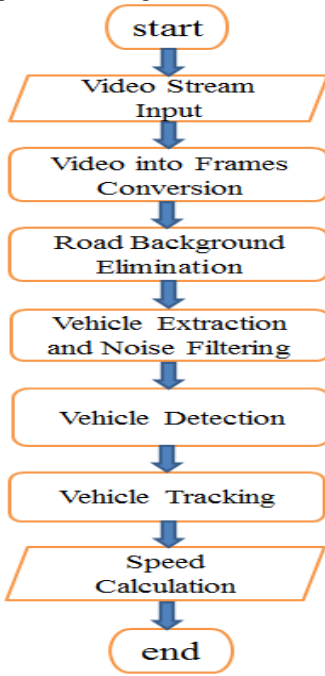

*Figure 2. Flowchart of the proposed method*

### A. Video Stream Input

First of all, a video is recorded with a camera fixed on road side. The embedded camera is with frame rate of 25 frames per second, where each frame with a resolution of 480×640 pixels. It should be noted that a camera with better frame rate and resolution at finer scale contains rich time and vehicle information.

### B. Video into frame conversion

From the captured video, all the frames are extracted individually using the simple image processing technique. The 30th frame of the video is shown in Fig. 3.

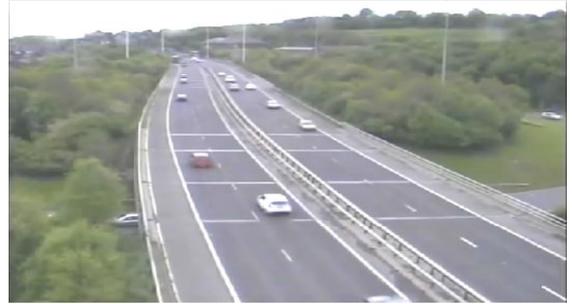

*Figure 3. Frame extracted (Fe)*

### C. Road background elimination

Video is a sequence of consecutive image frames. The numerical pixel values in each frame carry the background and the foreground information. The background information consists of stationary objects like trees on road sides, buildings, parked vehicles or any other stationary objects, the climatic conditions and specifications of day and night. The foreground information consists of moving objects like moving vehicles, pedestrians or any other moving object. Background extraction is the key part of the moving vehicle detection algorithm and helps in extracting foreground information. Different techniques have been followed in detection algorithms for removing background information, some of which are presented here. Al-Bailo et al., [9] suggested the use of Gaussian distribution for background estimation and image segmentation. In this work, we first create road mask from the road image in Figure 3 by employing the image processing technique. The generated road mask is shown in Figure 4.

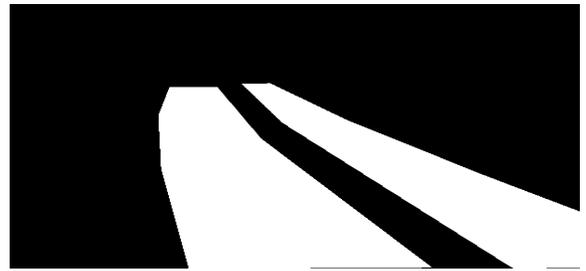

*Figure 4. Road Mask (Vp)*

Then, we multiply the video frames with road mask (Vp) image, as shown in equation 1. The resultant image without background is shown in Fig. 5. However, this step helps in removing all the stationary objects like trees, poles and sign boards. This step also helps to simplify the task and find the direction of the moving vehicle.

$$N(p) = F(e) \times V(p) \qquad (1)$$

Where;
- $F(e)$ is an image point value, shown in Fig. 3.
- $N(p)$ is a new image point.
- $V(p)$ is mask value for point p.

If a pixel value has been eliminated, V(p) = 0 and otherwise V(p) = 1.

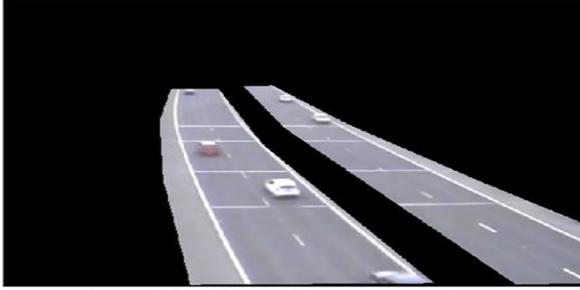

Figure 5. Road image without stationary objects

### D. Vehicle Extraction and Noise Filtering

After eliminating the static objects, the next step is to extract the vehicles from the road. For this we have to remove the background road from the image in Fig. 5. The resultant image is in Fig. 6.

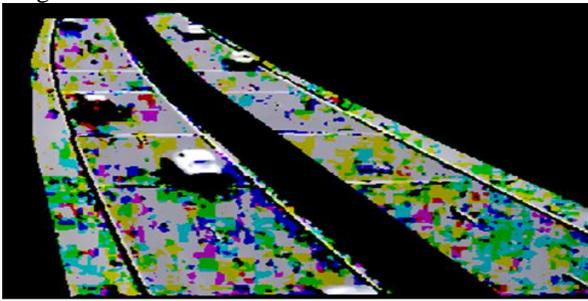

Figure 6. Simple subtracted Image.

It should be noted that the generated image is highly affected by noise. In order to resolve this problem, we apply threshold value to the preceding subtraction method, which not only helps in reducing noise effect but also gives a resultant image with finer detail. Different threshold values were tested. Lowering threshold value extracts all the information but does not completely eliminate noise in the resultant image. Increasing threshold value results in filtering noise while sacrificing the required information. The results of different threshold values are shown in Fig. 7(a, b, c, d) below.

Consequently, a threshold of 40 was found as an optimal value. A mathematical equation for thresholding is shown below.

$$\alpha(x,y) = \begin{cases} 0; & f(x,y) < 40 \\ 1; & f(x,y) \geq 40 \end{cases} \quad (2)$$

Fig. 7 shows more detailed information of vehicles as compared to the frame in Fig. 6.

However, threshold value helps in removing the majority of colored spots however, there might still be some spots left due to varying lighting or unusual weather. In order to remove these spots, we have used mean, mode and median filter. Mean filter removes the noise to a greater extent at the cost of blurring the image, which results in removing the far/small sized vehicles. Instead, we found median filter to be optimal for our problem. The smoothing from average filter causes blurring the information in the output image as shown in Fig. 9(a). On the other hand, median filter provides better results, as it preserves edges and suppresses salt and pepper noise as shown in Fig. 9 (b). The median filter works on reading each image pixel and its nearby neighbors and determines whether the corresponding pixel is a representative of its surrounding. The median filter replaces the corresponding pixel with the median of neighbor values as shown in Fig. 8.

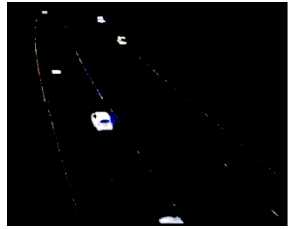 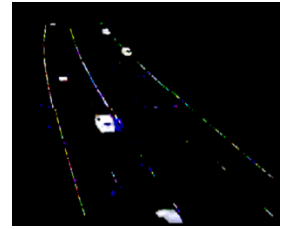

Figure 7 (a). Threshold = 10   Figure 7 (b). Threshold = 20

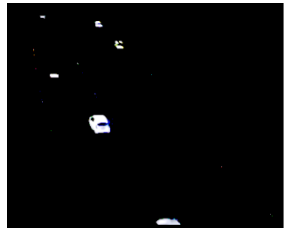 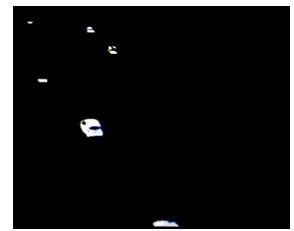

Figure 7 (c). Threshold = 30   Figure 7 (d). Threshold = 40

Figure 8. Median filtering picks the values at (x+1, y-1), (x+1, y), (x+1,y),..... and computes the result

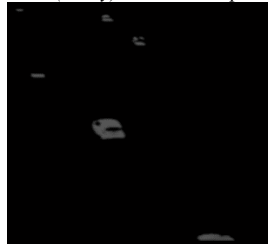 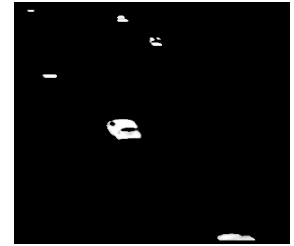

Figure 9(a). Average filter   Figure 9(b). Median Filter

Fig. 9 shows the processed image that is free of blobs and spots but still having effects of holes and cutting edges due to median filter.

In order to remove holes and cutting edges, we use morphological operations. Three main processes are applied namely; opening, closing and dilation. Opening and closing are intended to remove holes from the detected vehicle. The opening process is the dilation process after erosion. The dilation is the interaction of foreground pixel and structuring element. The resulting image is shown in Fig. 10.

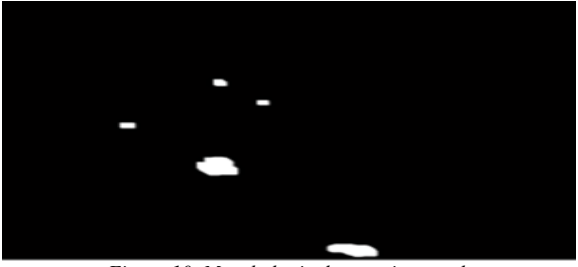

*Figure 10. Morphological operation result*

### E. Vehicle Detection

After suppressing noise from each frame, the next step is to locate the moving vehicle in the corresponding frame. This process is called vehicle detection. For moving object detection different techniques have been used, some of which are mentioned here. Kumar et al.,[10] used the concept of GPU for real time moving object detection. Ma et al.,[11] used the concept of Gaussian Mixture Model for moving object detection and Girshik et al.,[12] presented the concept of Rich Feature Hierarchies. However, in this work the vehicle detection is done by using very simple and easy method. We first detect contour of vehicles through edge detection algorithms based on mathematical operations of image derivatives. This actually works by detecting whether an edge passes through a given pixel. Steep gradient shows the detection of an edge but slow changes in gradient may generate opposite result. First derivative is useful in detecting the rate of intensity change in frames. While second derivative is used in determining the rate of maximum change of intensities in the image. We examined different detection algorithms based on Prewitt, Sobel, Laplacian and Canny filters. Prewitt produces better results on this particular task.

After that a random color to the first contour in the observation frame that is detected. Then the color of the primary vehicle is compared with the adjacent vehicle. If both the colors are same then the adjacent point is assigned a different color, otherwise no action is taken. This process is shown in Fig. 11.

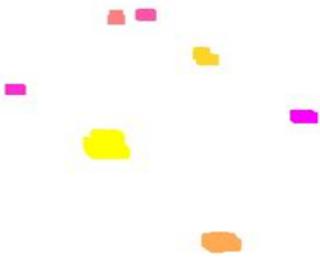

*Figure 11. Contour labeling.*

### F. Vehicle Tracking

Once the vehicle is detected, the next task is to find the track of the vehicle in frames. The geometric centers of the detected vehicle can be expressed as

$$X_c = \frac{\sum x_j}{n} \qquad (3)$$

$$Y_c = \frac{\sum y_j}{n}, \qquad (4)$$

where $X_c$ and $Y_c$ are center coordinates of the vehicle and $x_j$ and $y_j$ are coordinates of one of the $n$ images from the area limited to the external contour of the vehicle. The vehicle detected in first frame is considered as reference and it places a hole on the center of the vehicle. Then by using region growing method the circle radius is incremented by one till the vehicle is determined in the next frame. As soon as the vehicle is matched, the corresponding record is stored and tracking of other vehicle is started.

In order to track a vehicle in between all vehicle images from frame $n$ to frame $n+1$, we must calculate all of its distances in between the frames by using centered coordinates $(x_k, y_k)$. The distance can be calculated by using the following equations:

$$d_k = \sqrt{(x_k - x_c)^2 + (y_k - y_c)^2} \qquad (5)$$

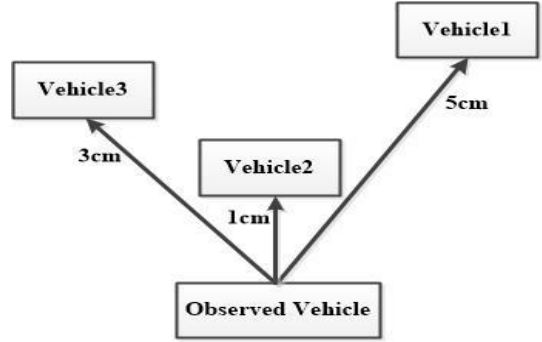

*Figure 12. Vehicle with shortest distance shows the corresponding vehicle in next frame.*

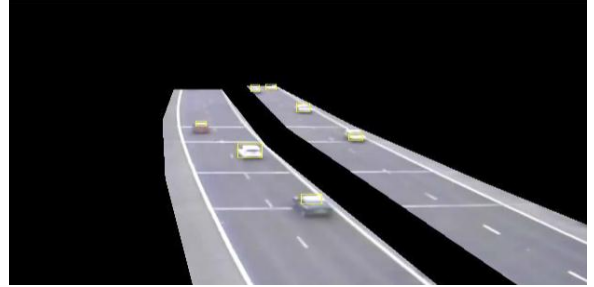

*Figure 13. Vehicle tracking in frames.*

After calculating the distances for each vehicle the min($d_k$) helps in tracking the vehicle in frame $n+1$. The shortest distance shows the corresponding vehicle in the next frame. In this way, all the vehicles can be tracked. See Fig. 12 and Fig. 13 for more clarification.

### G. Speed Calculation

After tracking the vehicle in the frame the last step is to calculate the vehicle speed and this can be done by using the speed formula that is give below; $s = v \times t$ (6)
Where "s" is equal to distance "d" calculated in previous step. and "t" depends upon video frame rate "k", that can be calculated as:

$$t = \frac{1}{k} \qquad (7)$$

Putting value of "t" in equation (6), the required speed formula can be written as;

$$v = ks \times v0 \qquad (8)$$

Where "v0" is the constant of proportionality depends upon the distance of vehicle from the camera. For simplicity and user convenience the speed of the vehicle is shown above the vehicle in a rectangular box as shown in Fig. 14.

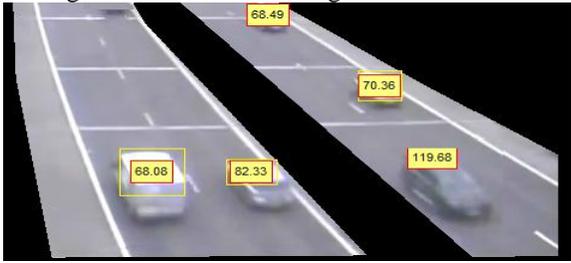

*Figure 14. Vehicle speed (Km/h)*

### III. CONCLUSION

This paper has presented an intelligent algorithm for speed monitoring. The method has been tested on different roads and provided efficient results. The proposed framework is simple, cost-effective and readily deployable. The algorithm only needs a camera fixed on the road attached to a computer for processing. The proposed framework helps in proper management of traffic flow resulting in limited chances of accidents. Furthermore, the framework can be used to detect and pull over the vehicles that violates traffic rules. The next task is to adapt the framework for road junctions. Moreover, the framework can be extended for working in tunnels to avoid any deadlock situation.